\documentclass[letterpaper, 10 pt, conference]{ieeeconf}  

\IEEEoverridecommandlockouts                              

\usepackage{graphics} 
\usepackage{epsfig} 
\usepackage{mathptmx} 
\usepackage{times} 
\usepackage{amsmath} 
\usepackage{amssymb}  
\usepackage{array}
\usepackage{subfigure}

\usepackage{mathtools}
\usepackage[justification=centering]{caption}
\title{\LARGE \bf
	Heterogeneous Vehicles Routing for Water Canal Damage Assessment
}

\author{Di Deng$^{1}$, Tao Pang$^{2}$, Prasanth Palli$^{1}$, Fang Shu$^{1}$ and Kenji Shimada$^{1}$
	\thanks{$^{1}$Faculty of Mechanical Engineering, Carniege Mellon University, 5000 Forbes Ave, Pittsburgh, PA 15213, USA.
		{\tt\small dengd@andrew.cmu.edu}}%
    \thanks {$^{2}$Computer Science and Artificial Intelligence Laboratory, Massachusetts Institute of Technology, 77 Massachusetts Avenue, Cambridge, MA 02139.
		{\tt\small pangtao@mit.edu}}%
}
\begin{document}
	
\maketitle
\thispagestyle{empty}
\pagestyle{empty}
\begin{abstract}
In Japan, inspection of irrigation water canals has been mostly conducted manually. However, the huge demand for more regular inspections as infrastructure ages, coupled with the limited time window available for inspection, has rendered manual inspection increasingly insufficient. With shortened inspection time and reduced labor cost, automated inspection using a combination of unmanned aerial vehicles (UAVs) and ground vehicles (cars) has emerged as an attractive alternative to manual inspection. In this paper, we propose a path planning framework that generates optimal plans for UAVs and cars to inspect water canals in a large agricultural area (tens of square kilometers). In addition to optimality, the paths need to satisfy several constraints, in order to guarantee UAV navigation safety and to abide by local traffic regulations. In the proposed framework, the canal and road networks are first modeled as two graphs, which are then partitioned into smaller subgraphs that can be covered by a given fleet of UAVs within one battery charge. The problem of finding optimal paths for both UAVs and cars on the graphs, subject to the constraints, is formulated as a mixed-integer quadratic program (MIQP). The proposed framework can also quickly generate new plans when a current plan is interrupted. The effectiveness of the proposed framework is validated by simulation results showing the successful generation of plans covering all given canal segments, and the ability to quickly revise the plan when conditions change.
\end{abstract}

\section{Introduction}
\begin{figure}[htp]
	\centering 
	\includegraphics[width=1.56in]{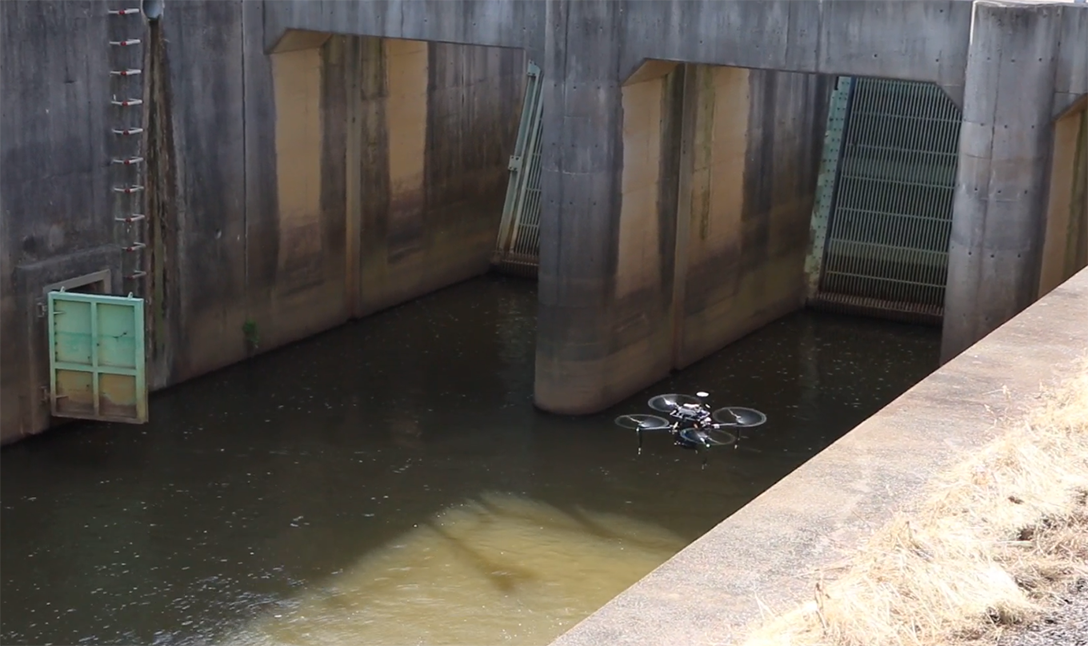}
    \includegraphics[width=1.74in]{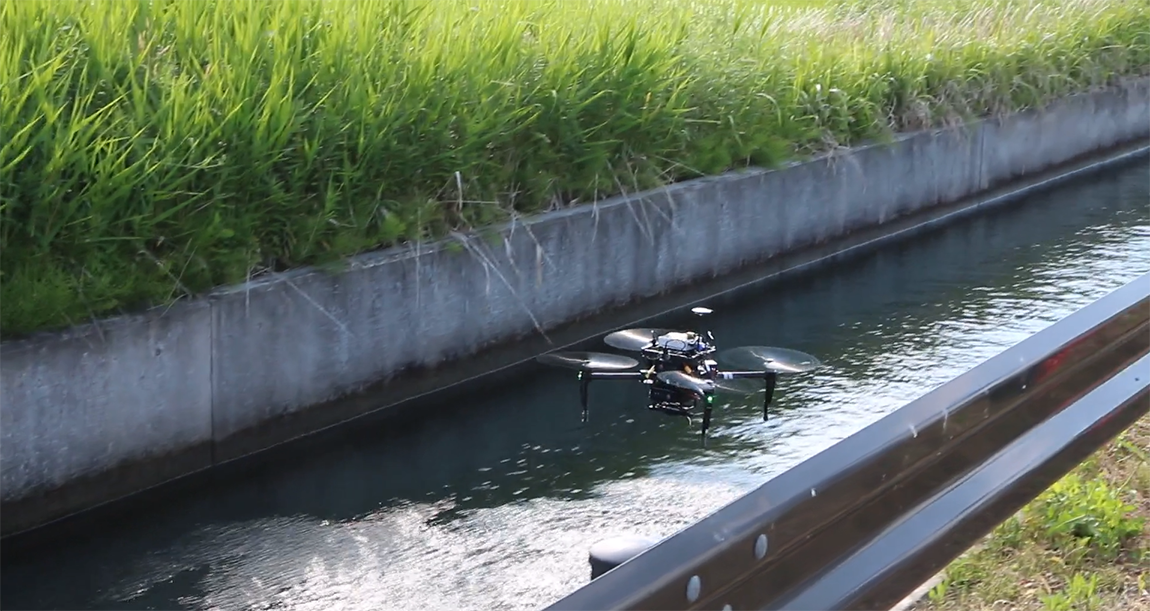}
	\caption{UAV taking photos of water canals for damage assessment}
	\label{fig:uav_inspection}
\end{figure}

In a typical Japanese agricultural town, the length of water canals extends to dozens of kilometers while the dry season during which inspection and repair can be conducted lasts only $1-2$ months per year. However, the current inspection process involves technicians walking along the canals and manually measuring and marking the damaged areas in a log book. Each technician can inspect only $0.5$ km a day. Furthermore, it takes over one month to convert the recorded data into digital information for guiding repairs. Water canal inspection using UAVs (Figure \ref{fig:uav_inspection}) would significantly reduce inspection time and labor cost by automatically identifying defects and registering their GPS coordinates. 

\begin{figure}[htp]
	\centering 
	\includegraphics[width=3.3in]{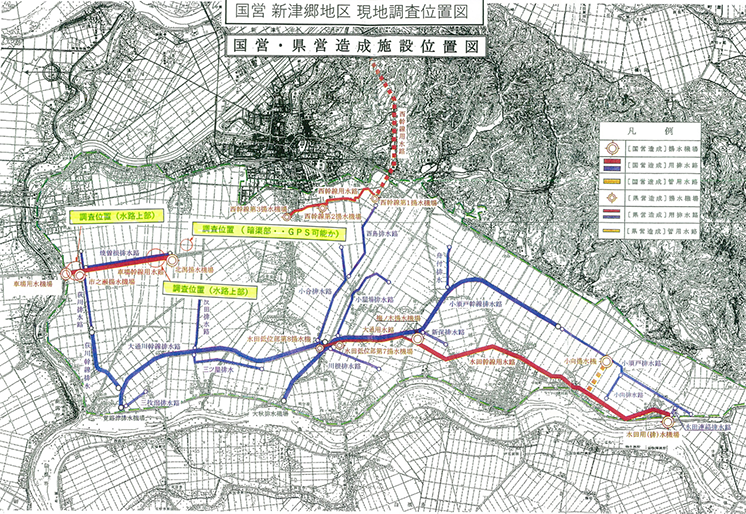}
	\caption{Water canal map of a town in Niigata ($10$ km $\times 7$ km)}
	\label{fig:map_of_water_canals}
\end{figure}

We will focus our efforts on a town in Niigata, an agricultural district along the northwest coast of Japan. In the area of interest shown in Figure \ref{fig:map_of_water_canals}, there are about $46.2$ km of water canals spread over tens of square kilometers of farmland. The scale of water canals is beyond the cruise and communication range of most commercially available UAVs, which can typically fly for $20-40$ minutes within $7$ km of the ground station. As a result, inspection of all water canals requires \textit{heterogeneous} vehicles, i.e. a combination of multiple UAVs and ground vehicles. 

A viable inspection strategy first divides all water canals into sub-regions that a UAV (or a fleet of UAVs) can cover within one battery charge. Once the UAVs complete inspection of one sub-region, they return to ground vehicles (cars driven by human operators) and are transported to the next sub-region while their batteries are swapped. 

There are several constraints that the UAVs and cars have to satisfy. First, as the cars are used as ground stations to monitor and recharge the UAVs, each UAV needs to have at least one car within its communication range at all time, and each UAV will return to a car when it runs out of battery. Second, as the cars travel on public roads, they have to abide by local traffic regulations and their speed is limited by real-time traffic conditions. Moreover, it is also necessary to quickly re-plan the paths should unexpected situations (such as UAV breakdown or change of traffic conditions) arise. 

This paper presents an MIQP-based planning framework that generates optimal inspection plans for a given fleet of UAVs and cars, accounting for the aforementioned constraints. The plan includes recharging and transportation between different regions. The proposed framework can also quickly generate feasible new plans if a current plan is interrupted. 

The structure of this paper is as follows: Section \ref{sec:problem_statement} describes the coverage planning problem we are trying to solve; Section \ref{ch:Algorithms} provides detailed descriptions of our proposed planning framework; simulation results are presented in Section \ref{ch:Result}; the conclusion and future work are stated in Section \ref{ch:Conclusion}. 

\section{Related Works}
Cooperative control of a multi-agent system reduces operation time, introduces redundancy and robustness to the system to better handle adversarial situations such as malfunction or failure of one or more agents. Araki \textit{et al.} built a robot swarm that can function as both ground and aerial vehicles, and developed planning methods using mixed integer program (MIP) \cite{araki2017multi}. Schillinger \textit{et al.} studied optimal planning algorithms for multiple ground robots using linear temporal logic \cite{schillinger2017multi}. Kim \textit{et al.} generates dynamically-feasible trajectories for multiple UAVs to cooperatively transport large objects \cite{kim2017motion}. 

Due to the expressiveness of integer variables, MIPs have been used to solve planning problems with complex constraints. Although current MIP solvers have worst-case exponential complexity \cite{bertsimas1997introduction}, MIPs can be solved fast enough for many nontrivial problems that are practically useful. Classical graph planning problems such as traveling salesman problem (TSP) \cite{hoffman2013traveling} and vehicle routing problem (VRP) \cite{laporte1992vehicle} can be written down and solved as mixed-integer linear programs (MILP). Avellar \textit{et al.} solves a minimum-time ground area coverage problem by converting it to a VRP \cite{avellar2015multi}. Dynamically-feasible, obstacle-free UAV trajectories can also be obtained by solving MIPs \cite{deits2015efficient,richards2002coordination}. Grotli \textit{et al.} studied UAV planning problems with fuel and communication constraints, but the communication ground stations are fixed \cite{grotli2012path}. Evers \textit{et al.} deals with the coverage problem and considers the uncertainty of fuel usage and weather conditions to provide a robust planning solution \cite{evers2014robust}. However, no refueling is planned, and only target nodes instead of edges are required to be visited. Lim \textit{et al.} covers both edges and nodes of a graph representing a power network using a fleet of UAVs and minimizes overall inspection time \cite{lim2016multi}. 

To the best of our knowledge, although a wide variety of planning problems has been solved with MIPs, the planning problem of heterogeneous vehicles, which allows multiple recharging/refueling of some of the vehicles and has constraints involving multiple types of vehicles, has yet to be addressed.  

\section{Problem statement \label{sec:problem_statement}}
\subsection{Graph representations of water canals and roads\label{sec:graph_generation}}
\begin{figure}[h]
	\centering
    \subfigure[Graphs of water canals with $78$ nodes and $77$ edges]	{\label{fig:graph_canal}\includegraphics[width=3.0in]{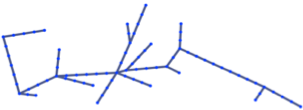}}
    \subfigure[Graphs of roads with $1391$ nodes and $1398$ edges]	{\label{fig:graph_road}\includegraphics[width=3.0in]{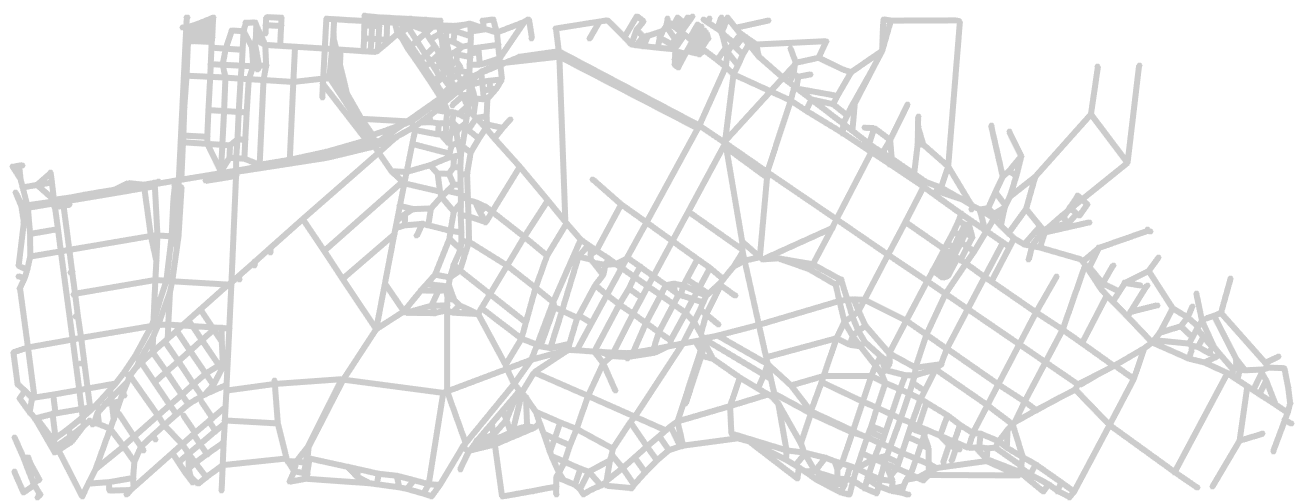}}
    \subfigure[Graph of water canals and roads]{\label{fig:graph_canal_road}\includegraphics[width=3.0in]{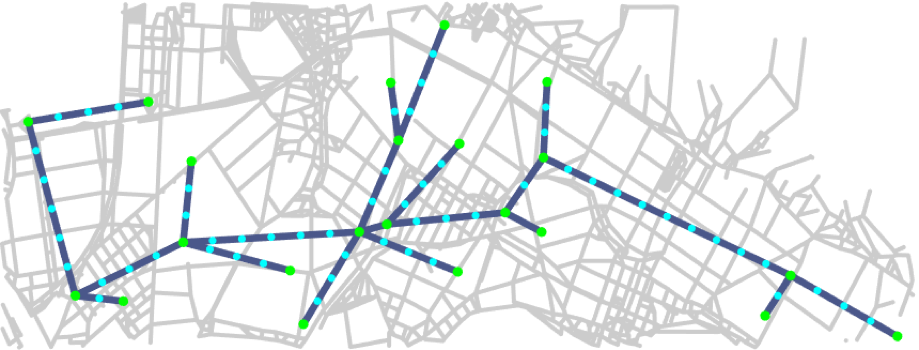}}
	\caption{Representation of water canals and roads}
	\label{fig:Graph_Generation}
\end{figure}

The first step in our planning framework is to abstract canal and road networks from a map into graphs (Figure \ref{fig:Graph_Generation}). GPS information of roads is extracted from the open-source map named OpenStreetMap \cite{OpenStreetMap} using OSMnx \cite{boeing2017osmnx}. NetworkX \cite{hagberg2008exploring} is used to assign nodes to end and intersection points on the road network and identify the adjacency matrix of the graph.

Water canals and roads are represented as two weighted graphs:
\begin{itemize}
\item $G_{canal}=(V_{canal},E_{canal})$, where $v \in V_{canal}$ is an intersection of or a point along the canals, and $e \in E_{canal}$ is a canal segment. The canal graphs are constructed in such a way that all edges in the graph have comparable weights, i.e. $weight(e_{canal}) \approx w_c$ for all $e_{canal} \in E_{canal}$, so that it takes a UAV the same amount of time to inspect all edges in $G_{canal}$. An important observation is that $G_{canal}$ is usually a tree;
\item $G_{road}=(V_{road},E_{road})$, where $v \in V_{road}$ is an intersection of roads, and $e \in E_{canal}$ a road segment.
\end{itemize}

\subsection{Heterogeneous planning problem \label{sec:coverage planning}}
\begin{figure}[h]
\centering
	\includegraphics[width=3.3in]{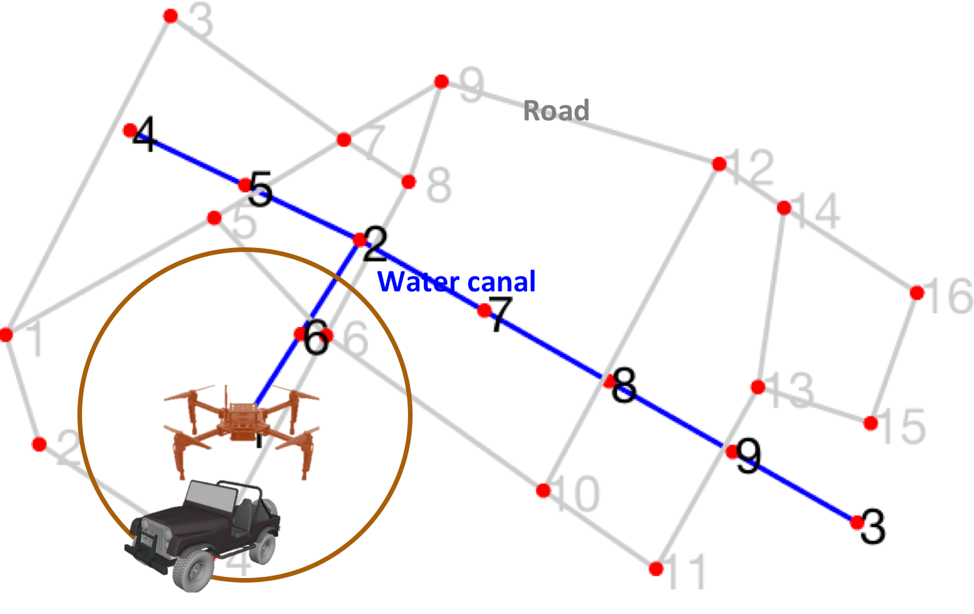}
	\caption{Every UAV needs a car within its transmission distance.}
	\label{fig:nearby}
\end{figure}

Given the following information:
\begin{itemize}
\item graphs of water canals and roads;
\item number of available cars and UAVs;
\item UAV battery life and transmission range;
\end{itemize}
we want to find paths for UAVs and cars that inspect all water canals as quickly as possible, subject to the following constraints:
\begin{itemize}
\item cars drive on the road graph $G_{road}$;
\item UAVs fly on the canal graph $G_{canal}$, except when taking off from and returning to the cars;
\item every UAV needs to have at least one car within its transmission distance, as shown in Figure \ref{fig:nearby};
\item every UAV takes off with a fully-charged battery, and needs to return to a car when its battery is depleted. 
\end{itemize}

\section{Proposed computational methods}\label{ch:Algorithms}
\subsection{Overview}
Using the canal and road graphs defined in Section \ref{sec:graph_generation}, the inspection paths for UAVs and cars are found through three steps: 
\subsubsection{graph partitioning} as the entire canal graph is usually much larger than what a UAV fleet can inspect with a single battery charge, the canal graph is partitioned into subgraphs that a given fleet can possibly inspect within the UAV's battery life ($20-40$ minutes). In addition, partitioning the canal graph significantly reduces the size and computation time of each of the path planning problems.

\subsubsection{heterogeneous vehicle path planning} for each subgraph generated from partitioning, the heterogeneous vehicle path planning algorithm generates paths for UAVs and cars that inspect all canals in the given subgraph, subject to the constraints given in Section \ref{sec:coverage planning}. 

\subsubsection{car routing between subgraphs} while being recharged, the UAVs are also being transported from one subgraph to another by cars. The car routing algorithm gives the minimum-length path that visits and inspects all canal subgraphs.

\subsection{Graph partitioning}
To minimize the time for swapping batteries and traveling between partitioned subgraphs, the objective is that partitions have comparable sizes. Let $K$ be the total number of UAVs and $M$ the number of edges a UAV can inspect with one fully charged battery. The water canal graph $G_{canal}$ consists of $N_n$ vertices and $N_e$ edges. Therefore, the maximum number of edges in a subgraph is $KM$, and an initial guess for the number of subgraphs is $s_o=\lceil\frac{N_e}{KM}\rceil$. 

The actual number of subgraphs is increased from $s_o$ until the following mixed-integer quadratic program (MIQP) becomes feasible. The MIQP of minimizing the traveling distance of UAVs has the following objective function and constraints:
\begin{equation}
\underset{x}{\mathrm{min.}}⁡\sum_{s=1}^S(\sum_{i=1}^N x_{si})^2,
\end{equation}
subject to:
\begin{equation}
\forall e,  \sum_{s=1}^S w_{se}=1,
\label{equ:coverage}
\end{equation}
\begin{equation}
\forall s,  K\leq\sum_{e=1}^E w_{se}\leq KM,
\label{equ:size}
\end{equation}
\begin{equation}
\forall s,  \sum_{e=1}^E w_{se}=\sum_{i=1}^N x_{si}-1, \mathrm{and}
\label{equ:connected}
\end{equation}
\begin{equation}
\forall s,x_{si}+x_{sj}-2w_{se}\geq0,
\label{equ:node_edge}
\end{equation}

where $x_{si}\in\{0,1\}$ represents whether Node $i$ is in Subgraph s, $w_{se}\in\{0,1\}$ represents whether Edge $e$ in Subgraph $s$. Constraint (\ref{equ:coverage}) requires that each edge belongs to one and only one subgraph. Constraint (\ref{equ:size}) sets $K$ and $KM$ as the lower and upper bounds of the number of edges in every subgraph. As a connected subgraph of a tree is also a tree, Constraint (\ref{equ:connected}) implies that each subgraph must be connected. Constraint (\ref{equ:node_edge}) shows that if Edge $e_{ij}$ is in Subregion s, both Node $i$ and Node $j$ are also in Subgraph $s$.

\begin{figure}[h]
	\centering
    \includegraphics[width=3.3in]{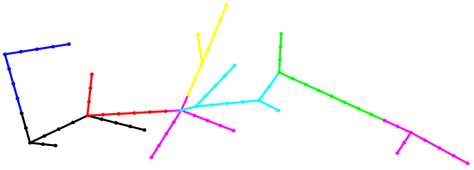}
	\caption{Graph partitioning result, different colors represent different subgraphs.}
	\label{fig:Graph_Partition}
\end{figure}

For example, given a fleet of 4 UAVs, in which every UAV can cover $3$ edges on a single battery charge ($K = 4$, $M = 3$), it is calculated that the graph of water canal should be partitioned into 8 subgraphs. The partitioning result is shown in Figure \ref{fig:Graph_Partition}.

\vspace{2mm}

\subsection{Heterogeneous vehicles path planning for subgraphs \label{sec:heterogeneous_planning}}
Given fixed numbers of UAVs and cars, the problem of generating paths for UAVs and cars to inspect a canal subgraph can be formulated as an MIQP with the objective of minimizing the traveling distance of UAVs. In addition to the constraints in Section \ref{sec:coverage planning}, we make the following assumptions:
\begin{itemize}
\item UAVs and cars only travel to adjacent nodes over one time step;
\item All canal edges in a subgraph must be inspected by a UAV;
\item It takes a UAV one time unit to inspect an edge in the canal subgraph; and
\item The time taken to fly between the canals and the vehicles at the beginning and end of the inspection is negligible compared to the inspection time. This is reasonable because the UAVs need to fly very slowly in a zig-zag pattern in order to acquire clear images of the canal walls. In comparison, the UAVs can fly reasonably fast when traveling between the canals and the cars.  
\end{itemize}

Let $N_{s}$ be the total number of edges in the canal subgraph indexed by $s$, the planning time horizon $T$ is increased from the initial guess $t_0=\lceil \frac{N_{s}}{KM}\rceil$ until the following MIQP becomes feasible. The MIQP of minimizing the traveling distance of UAVs has the following objective function:
\begin{equation}
\min_{x,y}\sum_{k=1}^K\sum_{t=1}^T\sum_{i=1}^{N_{canal}}(x_{k(t+1)i}-x_{kti})^2, 
\end{equation}
and constraints:
\begin{equation}
\forall e\in E_s,\sum_{k=1}^K\sum_{t=1}^T\sum_{d=1}^2w_{ektd}=1, 
\label{equ:coverage_constraint}
\end{equation}
$\forall k,t,e,$
\begin{equation}
x_{kti}+x_{k(t+1)j}-2w_{ekt1}\geq0,
\label{equ:node_edge_constraint1}
\end{equation}
\begin{equation}
x_{ktj}+x_{k(t+1)i}-2w_{ekt2}\geq0, 
\label{equ:node_edge_constraint}
\end{equation}
$\forall k,k_{car},t $
\begin{equation}
\left[\begin{array}{c}y_{k_{car}t1} \\ \vdots \\ y_{k_{car}tN_{road}}\end{array}\right]\leq R\left[\begin{array}{c}x_{kt1} \\ \vdots \\ x_{ktN_{canal}}\end{array}\right]+(1-\omega_{ktk_{car}})\left[\begin{array}{c}1 \\ \vdots \\1\end{array}\right],
\label{equ:communication_constraint1} 
\end{equation}
\begin{equation}
\forall t,k \sum_{k_{car}=1}^{K_{car}}\omega_{ktk_{car}}\geq1, 
\label{equ:communication_constraint} 
\end{equation}
\begin{equation}
\forall k,t,k_{car} \sum_{i=1}^{N_{canal}}x_{kti}=1,
\sum_{i=1}^{N_{road}}y_{k_{car}ti}=1, 
\label{equ:position_constraint}
\end{equation}
\begin{equation}
	\forall k,t, k_{car} \left[\begin{array}{c} x_{k(t+1)1}\\ \vdots \\ x_{k(t+1)N_{canal}}\end{array}\right]\leq A_{canal}\left[\begin{array}{c}x_{kt1}\\ \vdots \\ x_{ktN_{canal}}\end{array}\right] 
	\label{equ:constraint_connectivity_UAV},
	\end{equation}
	\begin{equation}
\left[\begin{array}{c} y_{k_{car}(t+1)1}\\ \vdots \\ y_{k_{car}(t+1)N_{road}}\end{array}\right]\leq A_{road}\left[\begin{array}{c}y_{k_{car}t1}\\ \vdots \\ y_{k_{car}tN_{road}}\end{array}\right],
	\label{equ:constraint_connectivity_car}
\end{equation}
where $x_{kti}\in \{ 0,1 \}$ represents whether UAV $k$ at time $t$ is at Node $i$ of the canal subgraph; $y_{kti} \in \{ 0,1 \}$ represents whether Car $k$ at time $t$ is at Node $i$ of the road subgraph; $\omega_{ktk_{car}}\in\{0,1\}$ represents whether UAV $k$ is within the communication range of Car $k_{car}$ at time $t$. $K_{car}$ is the total number of cars in the fleet. $R \in \mathbb{R}^{N_{road}\times N_{canal}}$ is the transmission matrix containing only $0$s and $1$s. $R(i,j) =1 $ if the Euclidean distance between Node $v_i \in V_{road}$ and Node $v_j \in V_{canal}$ is less than the UAV's maximum transmission distance. $A_{canal}$ and $A_{road}$, containing only $0$s and $1$s, are the adjacency matrices of the water canal and road subgraphs, respectively. $A(i,j)=1$ if node $i$ and node $j$ are adjacent.

Constraint (\ref{equ:coverage_constraint}) means that all edges must be inspected once. $w_{ektd} \in \{ 0,1 \}$ represents whether Edge $e$ is visited by the UAV $k$ starting at time $t$ in the direction $d$. For an edge $e_{ij} \quad (i<j)$, $d=1$ means that when the edge is inspected by an UAV, the UAV first visits Node $i$ and then Node $j$. Similarly, $d=2$ means that the UAV visits first Node $j$ and then $i$. $w_{ekt1}=1$ implies that $x_{kti}=1$, $x_{k(t+1)j}=1$, and $w_{ekt2}=1$ implies that $x_{ktj}=1$ and $x_{k(t+1)i}=1$. This relationship between $w_{ektd}$ and $x_{kti}$ is summarized in Constraint (\ref{equ:node_edge_constraint1}) and (\ref{equ:node_edge_constraint}). 

Constraints (\ref{equ:communication_constraint1}) and (\ref{equ:communication_constraint}) imply that for every UAV at all time, there must be at least one car within its transmission distance. Constraint (\ref{equ:position_constraint}) shows that at any time step, all cars and UAVs must appear at only one node. Constraints (\ref{equ:constraint_connectivity_UAV}) and (\ref{equ:constraint_connectivity_car}) require that UAVs and cars only travel to adjacent nodes between consecutive time steps. 

\begin{figure}[h]
	\centering
	\includegraphics[width=3.0in]{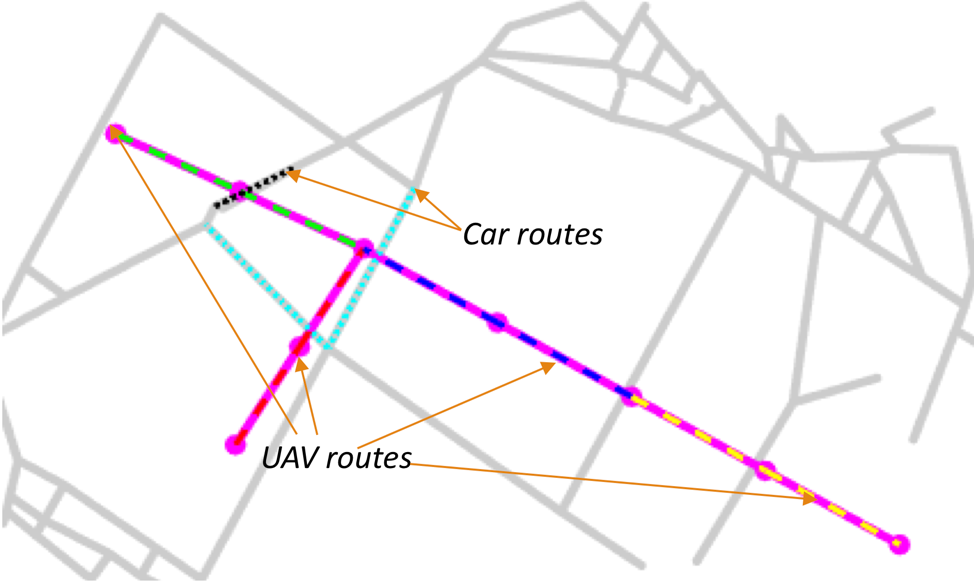}
	\caption{Planning result for one subgraph}
	\label{fig:plan_result}
\end{figure}

The planning result for one subgraph is shown in Figure \ref{fig:plan_result}. The magenta graph is a subgraph of $G_{canal}$ and the grey graph a subgraph of $G_{road}$. The fleet has 4 UAVs and 2 cars. The UAV inspection paths are shown on top of $G_{canal}$ as dashed lines with different colors. The car paths are shown as dotted lines on top of $G_{road}$.

\subsection{Re-planning}
When executing a plan on a subgraph, re-planning is necessary in the event of UAV failure or traffic jam. To re-plan at $t>0$, an MIQP similar to the one in Section \ref{sec:heterogeneous_planning} is constructed, with the following modifications:
\begin{itemize}
\item Canal edges already inspected at time $t$ are removed from the canal subgraph.
\item Congested roads are removed from the road subgraph. 
\item Battery life of UAVs is updated to the remaining battery life. 
\item All UAVs and cars start at their positions in the original plan at time $t$.
\end{itemize}

\begin{figure}[h]
	\captionsetup{justification=centering}
	\centering
    \subfigure[Original plan (4 UAVs and 2 cars)]{\label{fig:r1}\includegraphics[width=1.50in]{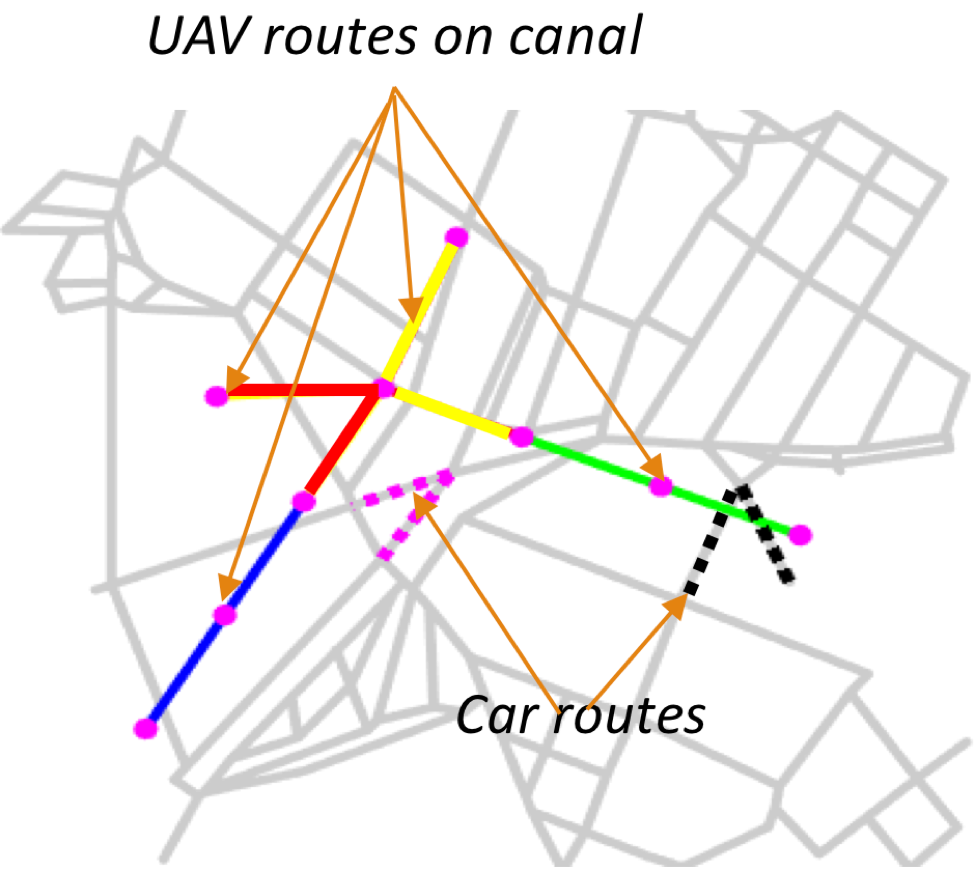}}
    \hspace{2mm}
    \subfigure[Re-planning at $t=1$. 4 edges have been inspected and removed from the canal subgraph.]{\label{fig:r2}\includegraphics[width=1.55in]{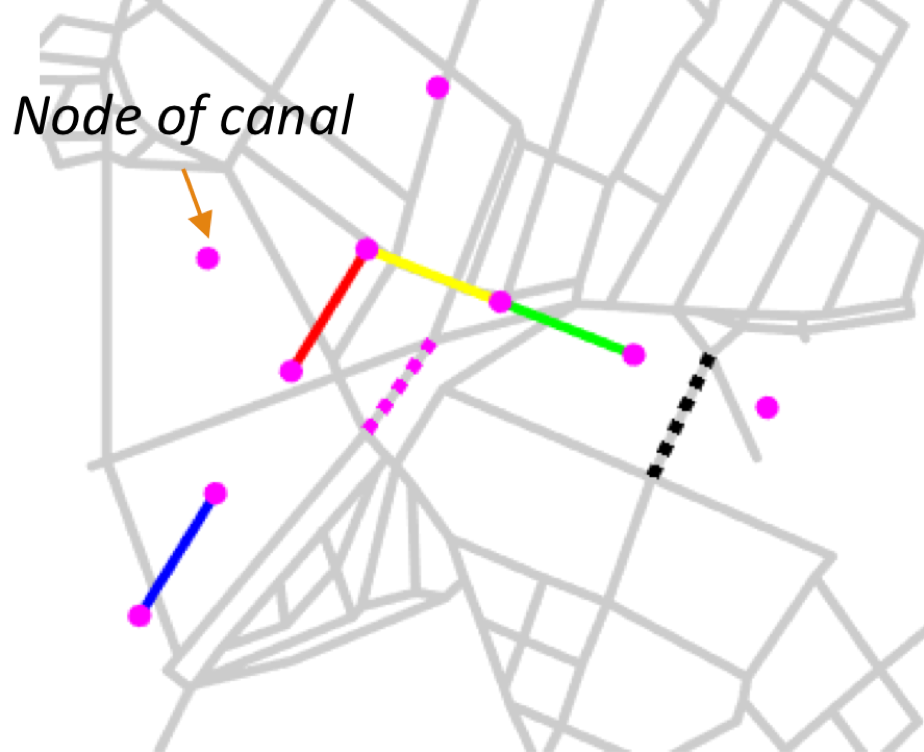}}
    \subfigure[Re-planning at $t=1$ due to traffic condition changes (some edges in the road graph are removed.)]{\label{fig:r3}\includegraphics[width=1.55in]{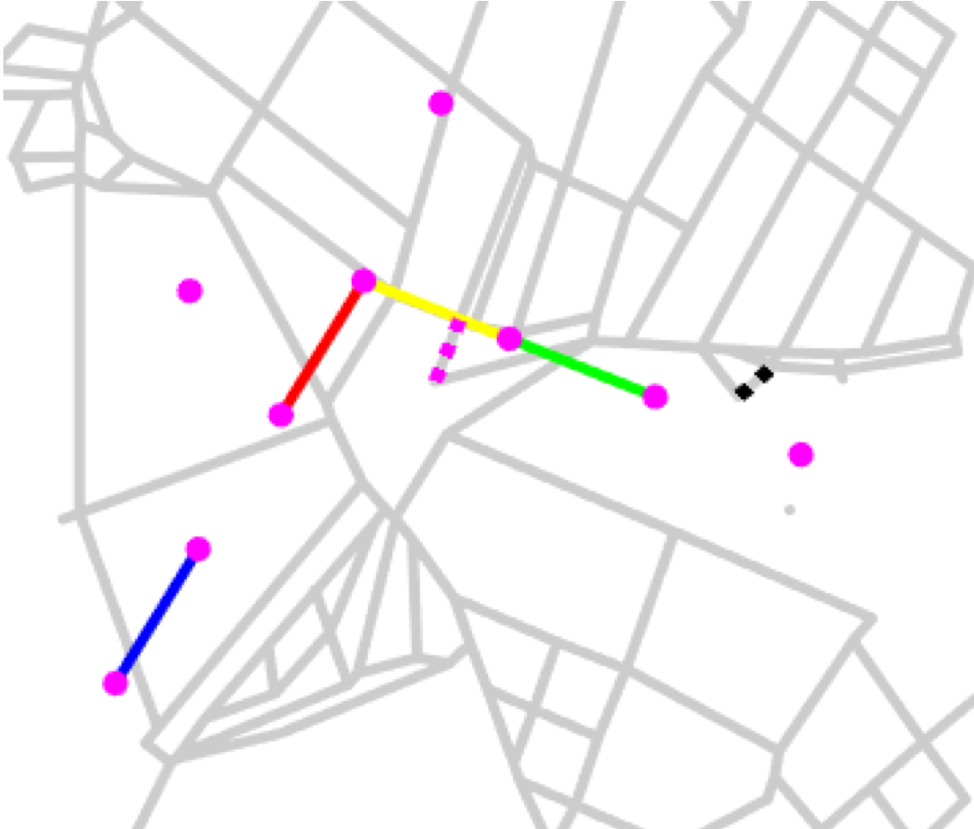}}
\hspace{2mm}
\subfigure[Re-planning at $t=1$ due to UAV failure (only 3 UAVs remain functioning)]{\label{fig:r4}
\includegraphics[width=1.55in]{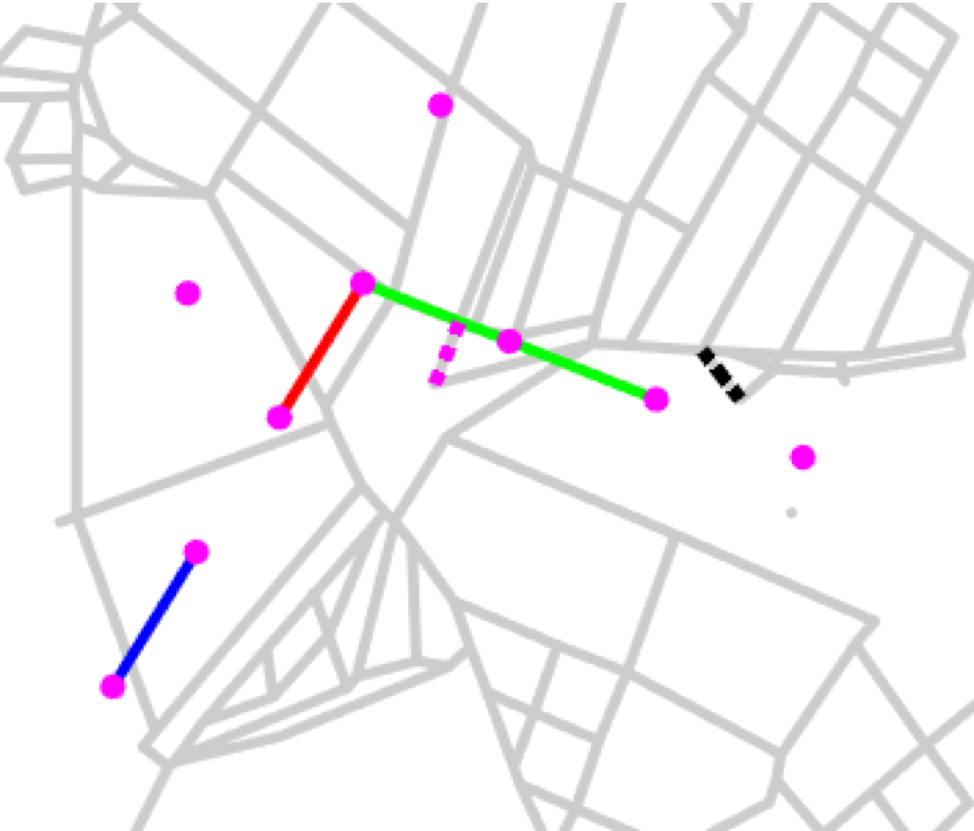}}
	\caption{Re-planning results}
	\label{fig:Replanning_fail}
\end{figure}

As shown in Figure \ref{fig:Replanning_fail}, the proposed re-planning method successfully generates new paths for UAVs and cars when a UAV fails or the traffic changes. 

\vspace{2mm}
\subsection{Car routing between subgraphs}
After finding an optimal plan to inspect a single subgraph, we need to find the shortest car route that visits all subgraphs. It is also required that the car route starts from and ends at a fixed location called the office. To plan car routes between canal subgraphs, a new graph $G_{subgraphs}$ whose nodes correspond to the canal subgraphs is constructed. As the cars are free to travel from any canal subgraph to any other canal subgraph, $G_{subgraphs}$ is fully connected but asymmetric (because of one-way roads). After determining the weights of the edges of the new fully-connected graph, the car routing problem, which searches for the shortest cycle that visits all nodes in $G_{subgraphs}$, can be solved as an Asymmetric Traveling Salesman Problem (ATSP).
 
\subsubsection{Determining edge weights in $G_{subgraphs}$}
Let $S$ be the total number of subgraphs generated from canal graph partitioning. The total number of nodes in $G_{subgraphs}$ is $S+1$ because the office is also included as a node. Let $Q \in R^{(S+1)\times(S+1)}$ be the matrix of edge weights of $G_{subgraphs}$, in which $Q_{AB}\coloneqq Q(A,B) $ is the weight of the directed edge pointing from Node $A$ to Node $B$. We want a $Q$ that minimizes the total travel distances of all cars when they carry UAVs from one subgraph to another (i.e. Subgraph A to Subgraph B). 

\begin{figure}[h]
\begin{center}
	\includegraphics[width=3.3in]{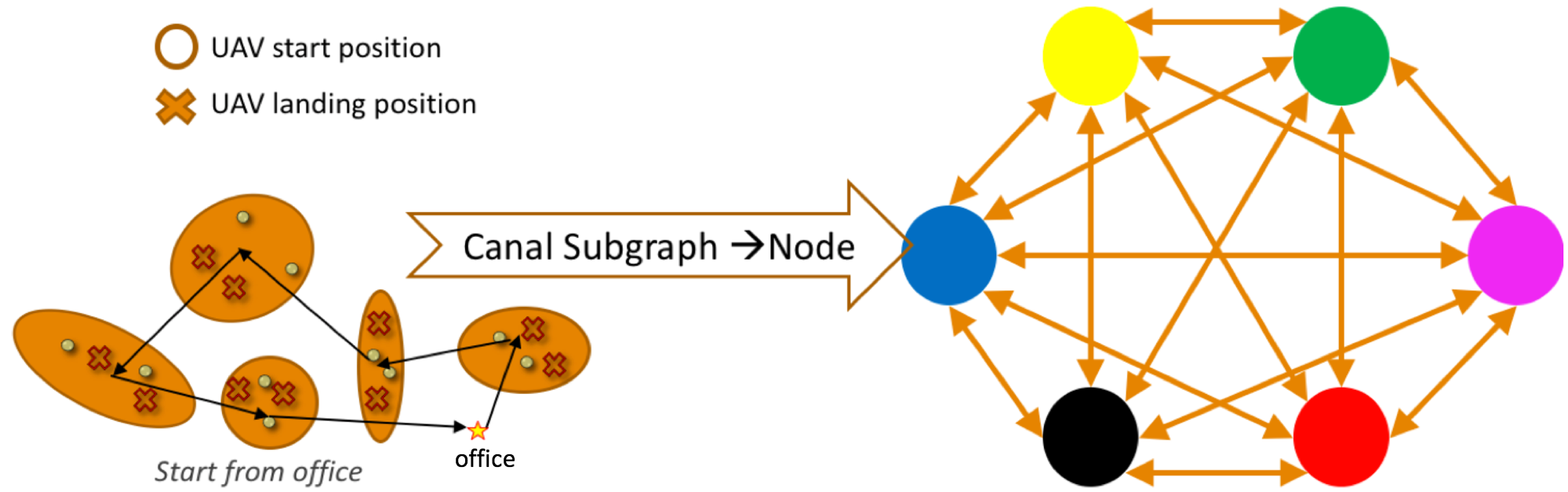}
	\caption{Transform into ATSP problem}
    \label{fig:atsp}
\end{center}
\end{figure}

In the inspection planning of subgraphs (Section \ref{sec:heterogeneous_planning}), the planner gives a starting and leaving node for every car in every subgraph. The UAVs take off at the starting nodes (shown as $\bigcirc$ in Figure \ref{fig:atsp}) and land at the leaving positions ($\times$ in Figure \ref{fig:atsp}). $Q_{AB}$ can be determined by finding the shortest of all paths connecting a leaving node ($\times$) of Subgraph A and a starting node ($\bigcirc$) in Subgraph B. This can be formulated as the following MIQP:

\begin{equation}
Q_{AB}=\min\sum_{i=1}^{K_{car}}\sum_{j=1}^{K_{car}}d_{ij}x_{ij},
\end{equation}
subject to:
\begin{equation}
	\forall i, \sum_{j=1}^{K_{car}}x_{ij}=1,
	\forall j, \sum_{i=1}^{K_{car}}x_{ij}=1,
    \label{equ:complete}
\end{equation}
where $d_{ij}$ is the length of the shortest path between Node $i$ and Node $j$ in the road graph, and $x_{ij} \in \{0,1\}$ is whether car travels from Node $i$ in Subgraph $A$ to Node $j$ in Subgraph $B$. Constraint (\ref{equ:complete}) says that all ending nodes of Subgraph $A$ must be visited only once and that all starting nodes of Subgraph $B$ must be visited only once.

\subsubsection{ATSP}
\begin{figure}[h]
	\centering
    \captionsetup{justification=centering}
	\subfigure[Start and end nodes on graph]{\label{fig:original}\includegraphics[width=1.65in]{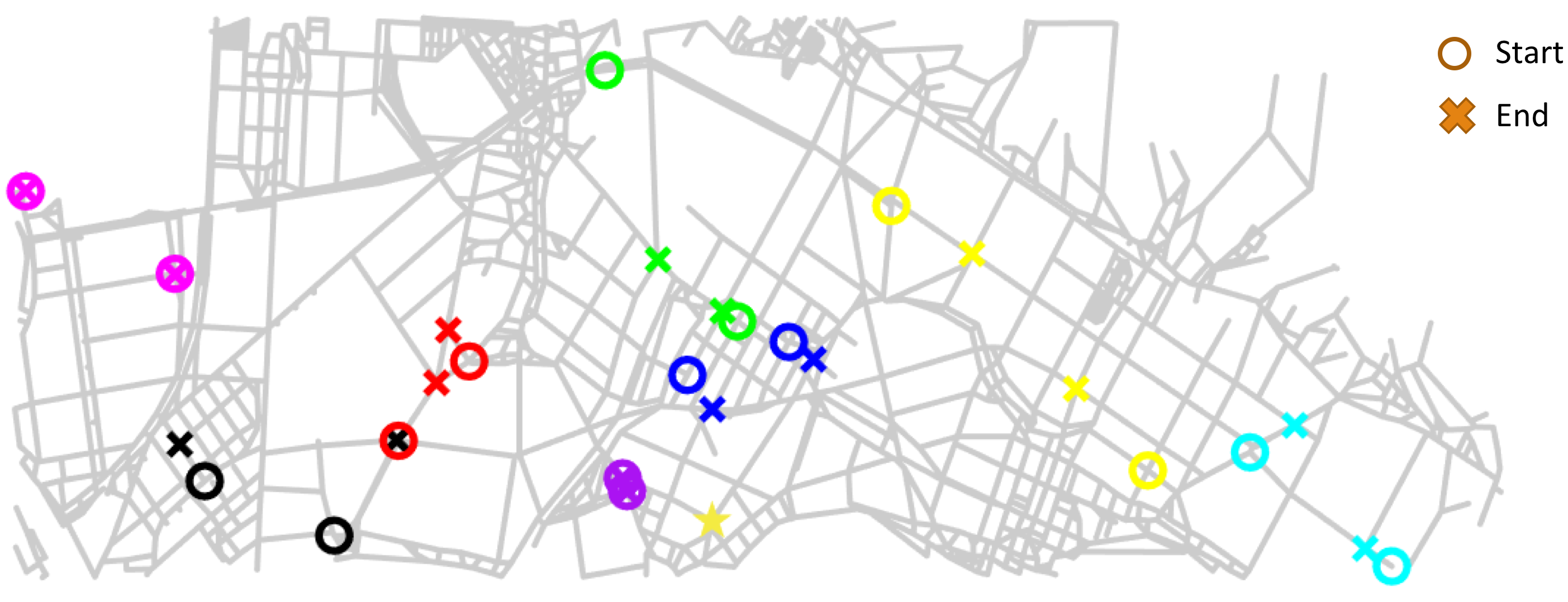}}
    \subfigure[Start and end nodes]{\label{fig:start_end}\includegraphics[width=1.65in]{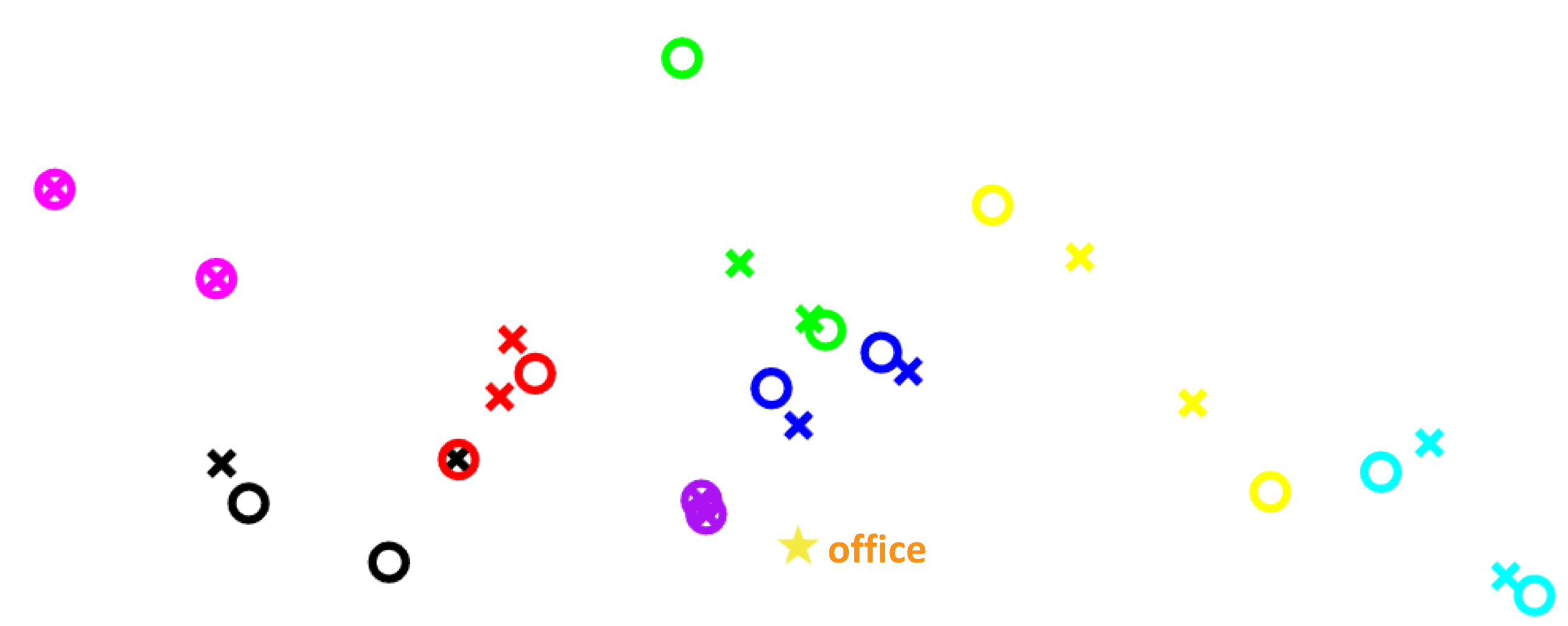}}
    \subfigure[Car-route between 2 subgraphs]{\label{fig:short2path}\includegraphics[width=1.65in]{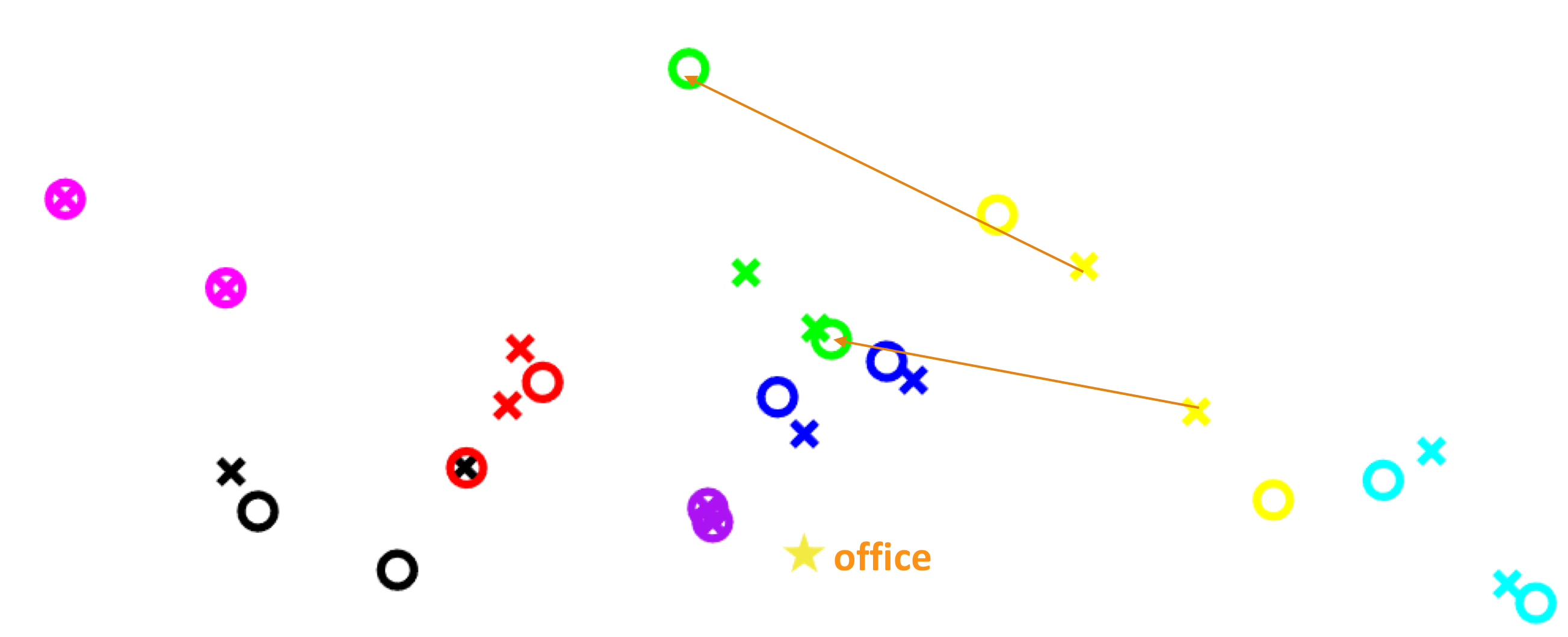}}
    \subfigure[ATSP problem]{\label{fig:subgraphs}\includegraphics[width=1.65in]{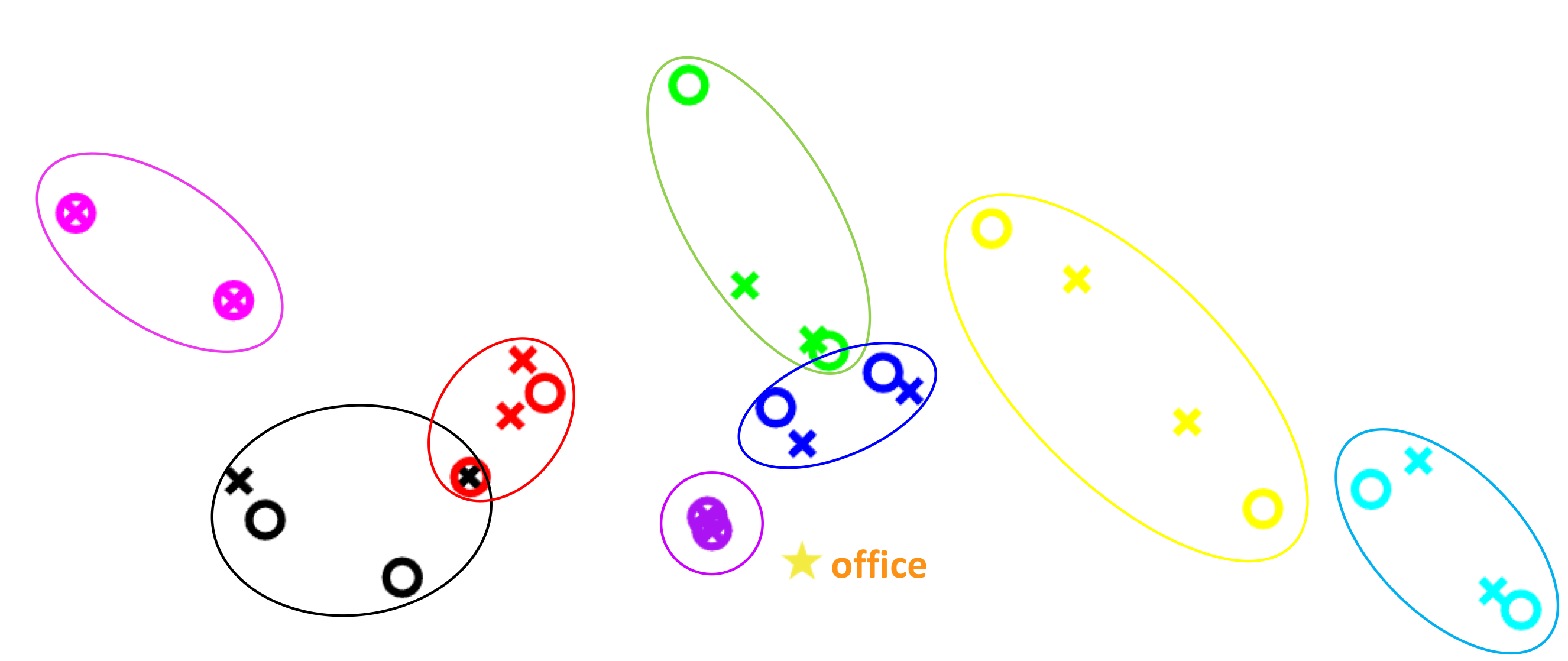}}
    \subfigure[ATSP result]{\label{fig:short_path}\includegraphics[width=1.65in]{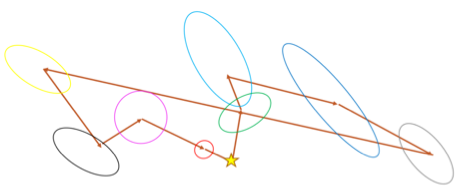}}
    \subfigure[Car routing result]{\label{fig:car_result}\includegraphics[width=1.65in]{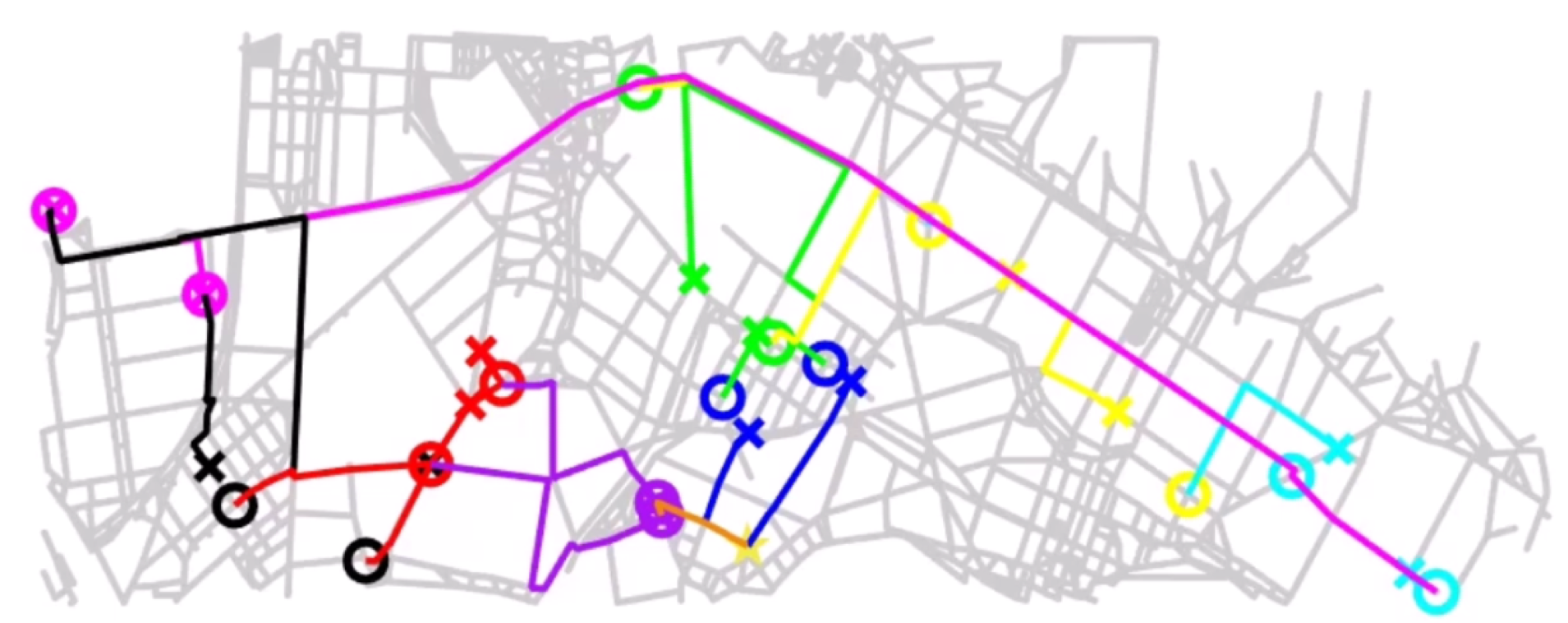}}
	\caption{Car routing between subgraphs}
	\label{fig:Car_Routing_between_Subgraphs}
\end{figure}

After finding $Q$, the car routing problem is reduced to an ATSP for a complete directed graph. The objective is to find the shortest closed path that starts from the office and visits all subgraphs: 
\begin{equation}
\min_x\sum_{t=0}^{S+1}\begin{bmatrix}x_{0t} & \dots & x_{(S+1)t} 	\end{bmatrix} Q	\begin{bmatrix}x_{0(t+1)} \\ \vdots \\ x_{(S+1)(t+1)}	\end{bmatrix},
\end{equation}
subject to:
\begin{equation}
x_{00}=1,x_{(S+1)(S+1)}=1, and
\label{equ:start_end_pos}
\end{equation}
\begin{equation}
	t\in\{1,\dots,S\}, \sum_{s=1}^S x_{st}=1,
	s\in\{1,\dots,S\}, \sum_{t=1}^S x_{st}=1,
    \label{equ:sub_visit_cons}
\end{equation}
where $x_{st}\in\{0,1\}$ denotes whether subgraph s is visited at time step $t$. Constraint (\ref{equ:start_end_pos}) says that cars must begin and end the journey from/at the office. Constraint (\ref{equ:sub_visit_cons}) means at the same time, only one subgraph is visited, and each subgraph can be visited only once. Figure \ref{fig:Car_Routing_between_Subgraphs} shows the shortest paths for cars.

\section{Results and Discussion}	\label{ch:Result}
The computation time of all methods in our inspection planning framework is benchmarked and summarized in Table \ref{table_bench}. The problem used for benchmarking considers the inspection of a water canal graph with $78$ nodes and $77$ edges. The inspection is conducted by a fleet consisting of $4$ UAVs and $2$ cars. Every UAV can cover 3 edges with a fully-charged battery. The MIPs are solved using GUROBI\cite{gurobi} on a laptop with Intel® Core™ i7-3520M CPU (dual core four threads).

\begin{table}[h]
	\caption{Benchmarking results}
	\label{table_bench}
	\centering
		\begin{tabular}{|c||c||c||c|}
			\hline
				Graph& Heterogeneous planning& Car &\\ partition	&  for subgraphs	&  routing & Total \\
			\hline
			0.02s&	412.22s	& 5.22s&	417.26s \\
			\hline
		\end{tabular}
\end{table}

\begin{figure}[h]
	\begin{center}
	\includegraphics[width=3.3in]{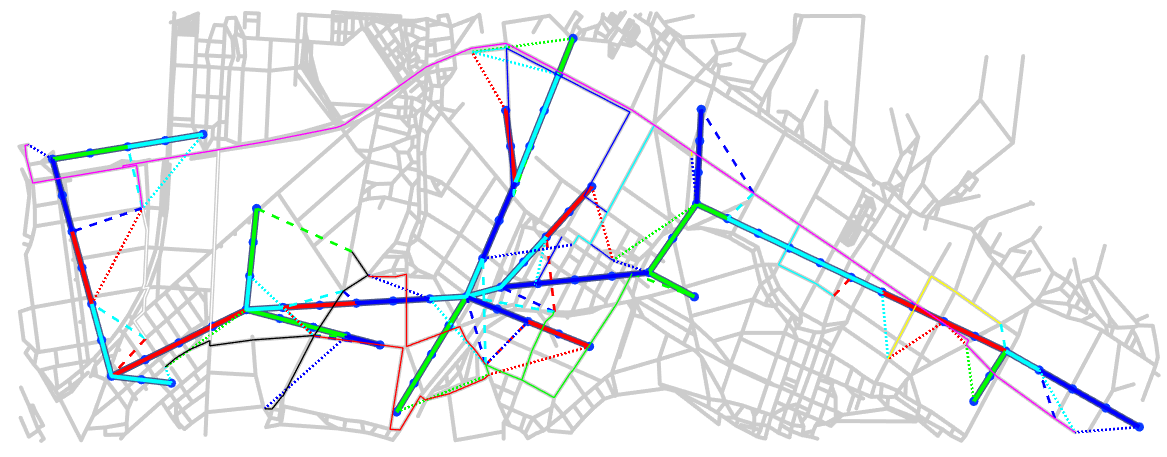}
	\caption{Planning result for 4 UAVs and 2 Cars}  
	\label{fig:overall_result}
    \end{center}
\end{figure}

\begin{figure}[h]
	\begin{center}
	\includegraphics[width=3.0in]{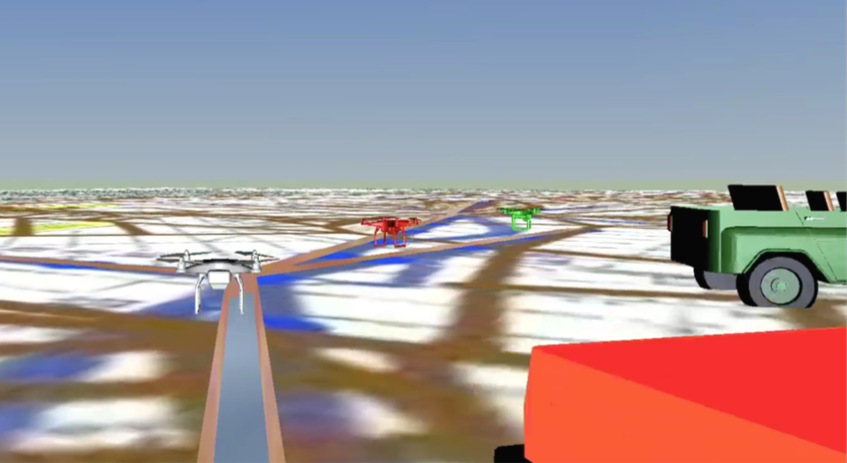}
	\caption{Simulation environment}
    \label{fig:3dSimulation}
   \end{center}	
\end{figure}

The planning result shown in Figure \ref{fig:overall_result} illustrates that our algorithm ensures the coverage of water canals and solves the problem caused by limited transmission distance and UAV battery life. Colored lines on graph of water canal represent the path of UAVs and colored lines on graph of road denote path of cars. Dotted lines between graph of water canal and graph of road are path of UAVs between cars and water canal. The entire inspection plan involving all cars and UAVs is also simulated in Simulink, as shown in Figure \ref{fig:3dSimulation}.

\begin{figure}[h]
	\begin{center}
    \subfigure[CAD model of a water canal subgraph]{\label{fig:cad}\includegraphics[width=3.3in]{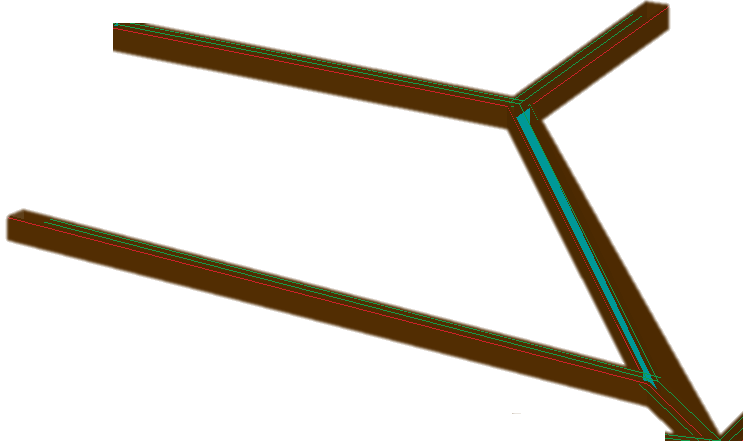}}
	\subfigure[Octomap reconstruction from simulated on-board stereo-camera]{\label{fig:recons}\includegraphics[width=3.3in]{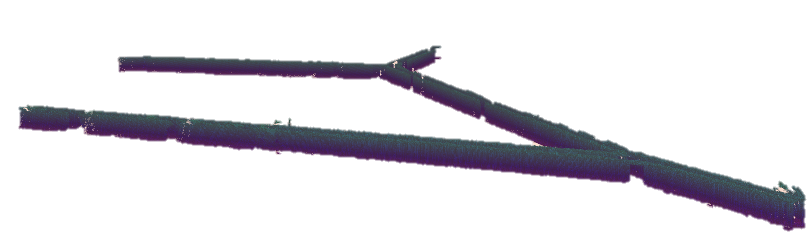}}
	\caption{Water canal subgraph reconstruction}
	\label{fig:reconstruction}
    \end{center}
\end{figure}

A 3D reconstruction of a water canal subgraph using Octomap \cite{hornung2013octomap} (Figure \ref{fig:recons}) is generated by simulating a UAV with stereo-camera flying inside the CAD model of water canals (Figure \ref{fig:cad}) following a planned path. The local octomap for all the UAVs are obtained from the disparity depth-map of a virtual stereo camera sensor in Gazebo, a realistic physics simulator. These local octomap are stitched using the ground truth poses from the simulator to obtain a unified octomap of the model. Figure \ref{fig:reconstruction} proves that our method inspects both walls of water canals and is expected to work effectively in real world scenarios.  

\begin{figure}[h]
	\begin{center}
	\includegraphics[width=3.3in]{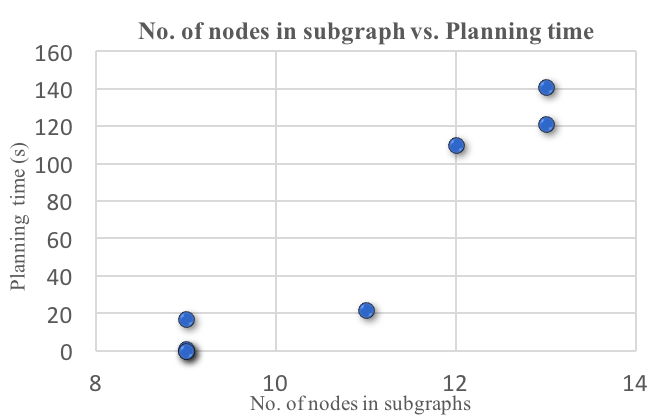}
    \caption{Planning time for all subgraphs}
    \label{fig:solving_time_}
    \end{center}
\end{figure}

It is shown in Figure \ref{fig:solving_time_} that the maximum planning time for canal subgraphs (algorithm in Section \ref{sec:heterogeneous_planning}) is 140, which can be further reduced by parallelization. Based on experiments, the time needed by cars to travel from one subgraph to another is about 10 minutes. If a new inspection plan is needed for the next subgraph due to different traffic conditions or the malfunctioning of some of the UAVs, the new plan will be ready before arriving at the next subgraph. 

It has been assumed in Section \ref{sec:heterogeneous_planning} that the time needed by one UAV to inspect one edge in $G_{canal}$ is 1 unit time, which is 10 minutes based on experiments. If we further assume that the time taken by cars to travel from one subgraph to another is 10 minutes, then the total inspection time for UAV-car fleets of different sizes can be calculated, and is summarized in Table \ref{table_time}. There is a 75\% reduction in inspection time by increasing the fleet size from 1 UAV and 1 car to 4 UAVs and 2 cars.

Based on experimental results, the efficiencies of different inspection methods are summarized in Table \ref{table_inspection_comparison}. Manual inspection involves a technician walking along the canals and looking for defects. In manual navigation, an operator flies a UAV manually along the canals, and another inspector looks for defects from the UAV's video feed. In our approach (heterogeneous vehicle routing), the UAV flies autonomously, records video of the canal walls, which are then analyzed with computer vision algorithms. Compared with current manual inspection and manual UAV flying method, the proposed heterogeneous vehicle planning framework using $4$ UAVs and $2$ cars would reduce inspection time by $90\%$ and $60\%$, respectively. These results could be improved further by using more UAVs and cars.

\begin{table}[h]
	\caption{Fleet size and inspection Time}
	\label{table_time}
	\centering
		\begin{tabular}{|c||c||c||c|}
			\hline
			Fleet size	&1 UAV 1 car	&3 UAVs 3 cars&	4 UAVs 2 cars\\
			\hline
			Inspection time & & &\\
            (minutes) &1320&	490	&330 \\
			\hline
		\end{tabular}
\end{table}

\begin{table}[h]
	\caption{Comparisons of different inspection methods}
	\label{table_inspection_comparison}
	\centering
		\begin{tabular}{|c||c||c||c|}
			\hline
				&Manual& Manual&	Heterogeneous \\
                &inspection & navigation& vehicle routing\\
			\hline
            Water canal& & &\\inspection time & & &\\($s/m^2$)
             &4.11 & 1.35 & 0.42\\
			\hline
Resolution	&NA	&0.2 mm-25 mm&	0.2 mm\\
			\hline
		\end{tabular}
\end{table}

\section{Conclusion}\label{ch:Conclusion}

Using UAV and car combinations to inspect large area of water canal reduces inspection time. A innovative algorithm is formulated to solve the new heterogeneous planning problem resulted from the UAV and car combinations. For the current map of water canal in an agricultural town in Japan, the planning algorithm runs online. The planning time is $140$ seconds for each subgraphs and re-planning time is less than $3$ seconds. Compared with manual inspection, using $4$ UAVs and $2$ cars can be expected to reduce inspection time by at most $90\%$. 

In the future, collision avoidance between vehicles, aerial constraints (weather, flying zones, busy aerial traffic, aerial regulations, etc.), ground traffic regulations (one-way, non-parking zones, speed limits, etc.) and uncertainty in maps and measurements of UAV/car localization will be added to the system to increase safety during inspection flights.

\section*{ACKNOWLEDGMENT}
The authors would like to thank TOPRISE Co., LTD. for their sponsor for this work and 
Mayuko Nemoto for help on problem formulation and running experiments.
\bibliographystyle{IEEEtran}
\bibliography{ref}
\end{document}